\documentclass[letterpaper, 10 pt, conference]{ieeeconf}  

\IEEEoverridecommandlockouts                              

\usepackage{bm}
\usepackage{outlines}
\usepackage{xspace}
\usepackage{xstring}

\usepackage{arydshln}
\usepackage{floatrow}
\usepackage{amsmath}
\usepackage{hyperref}
\usepackage{caption}
\usepackage{xcolor}
\usepackage{titlesec}
\usepackage{adjustbox}
\usepackage{amssymb}
\usepackage[disable]{todonotes}
\usepackage{subcaption}
\usepackage{graphicx}
\usepackage{cite}
\usepackage{tcolorbox}
\newcommand{\trajlen}{T_{\tau}}
\newcommand{\ed}[2]{#2}
\newcommand{\trajset}{\bm{\mathcal{T}}}
\newcommand{\tpr}{TPR\xspace}
\newcommand{\tnr}{TNR\xspace}

\newcommand{\dmax}{d_{\mathrm{max}}}
\definecolor{lightpurple}{RGB}{0,0,128}

\title{\LARGE \bf
Task-Oriented Active Learning of Model Preconditions
\\for Inaccurate Dynamics Models
}
\author{Alex LaGrassa$^{1}$, Moonyoung Lee$^{1}$, Oliver Kroemer$^{1}$
\thanks{A, LaGrassa, M. Lee, O. Kroemer are with Carnegie Mellon University Robotics Institute, Pittsburgh PA, USA \texttt{\{alagrass, moonyoul, okroemer,\}@andrew.cmu.edu}}
\thanks{This work was supported by NSF Grants No. CMMI-1925130 and IIS-1956163, ARL Grant No. W911NF-18-2-0218 as part of the A2I2 Program, and NSF/USDA NIFA AIIRA AI Research Institute 2021-67021-35329.}}

\begin{document}

\maketitle
\thispagestyle{empty}
\pagestyle{empty}

\begin{abstract}
     When planning with an inaccurate dynamics model, a practical strategy is to restrict planning to regions of state-action space where the model is accurate: also known as a \textit{model precondition}. 
     Empirical real-world trajectory data is valuable for defining data-driven model preconditions regardless of the model form (analytical, simulator, learned, etc...). However, real-world data is often expensive and dangerous to collect.
    In order to achieve data efficiency, this paper presents an algorithm for actively selecting trajectories to learn a model precondition for an inaccurate pre-specified dynamics model. Our proposed techniques address challenges arising from the sequential nature of trajectories, and potential benefit of prioritizing task-relevant data. 
    The experimental analysis shows how algorithmic properties affect performance in three planning scenarios: icy gridworld, simulated plant watering, and real-world plant watering. Results demonstrate an improvement of approximately 80\% after only four real-world trajectories when using our proposed techniques.
    More material can be found on our project website: \textcolor{lightpurple}{\href{https://sites.google.com/view/active-mde}{https://sites.google.com/view/active-mde}}.
\end{abstract}
\section{Introduction}

Many planning and control frameworks used in robotics rely on dynamics models, whether analytical or learned, to reason about how the robot's actions affect the state of the environment \cite{bates2019modeling, pfaff2020learning}. However, robots deployed in the real world frequently encounter unfamiliar environments and complex interactions, e.g., with deformable objects, where assumptions of simplified dynamics break, making the models deviate from reality. 
In such situations, previous works \cite{UnreliableMitrano2021, UnreliableDale2019, power2021keep, lagrassa2021learning} show how predicting model deviation with a Model Deviation Estimator (MDE) can be a powerful tool to restrict model-based planners to planning in regions of state-action space where the model is reliable, which we call \textit{model preconditions.}
Although using model preconditions defined by MDEs can improve the reliability of plans computed with inaccurate models, MDEs require real-world data, which can be expensive or dangerous, e.g., robot welding or pouring water, and drastically increase the cost of exploration. 

\begin{figure}[t]
    \centering
    \includegraphics[width=0.95\textwidth]{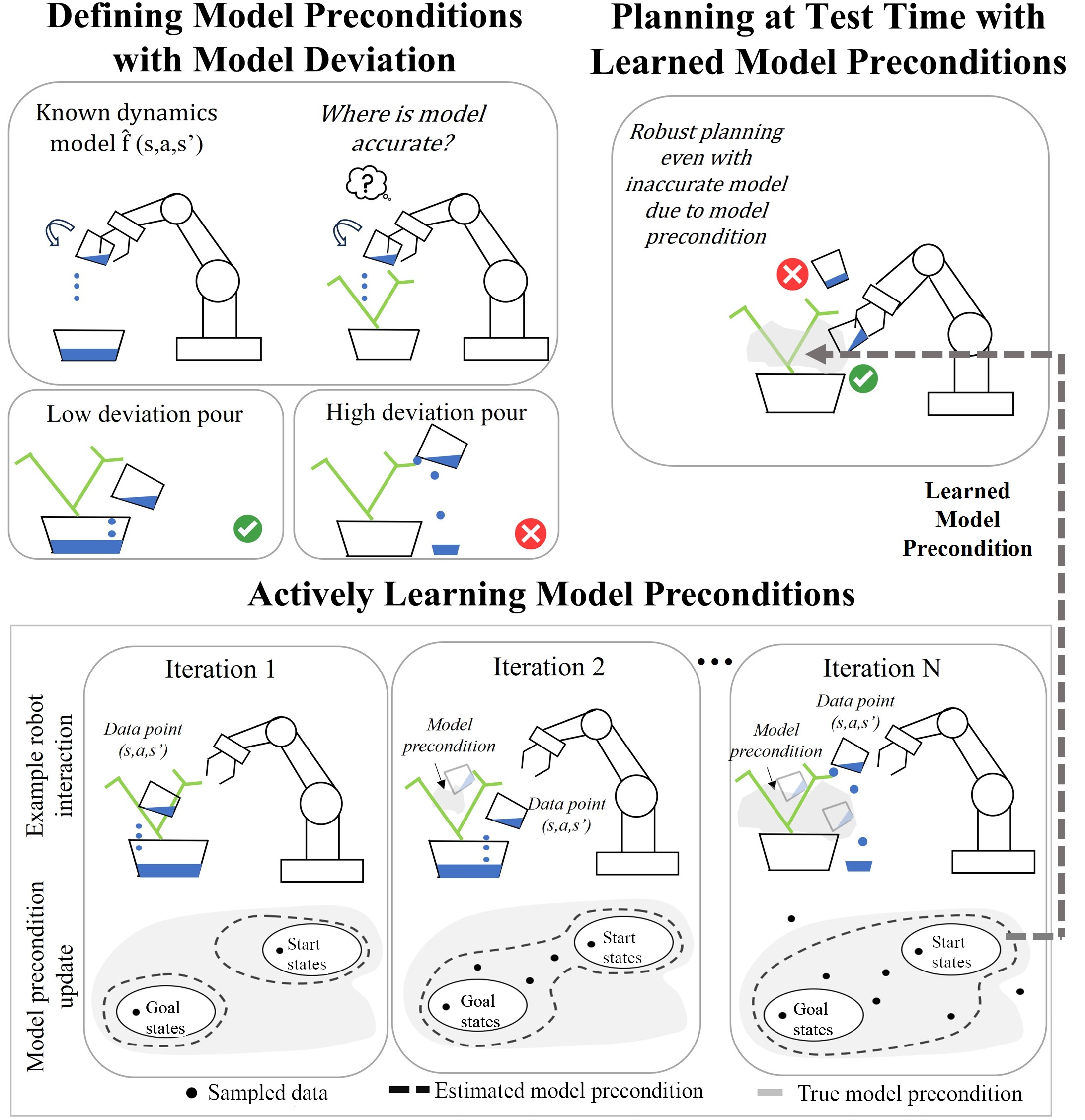}
    \vspace{-5pt}
    \caption{Illustrative example of using a planner and acquisition function to iteratively select informative trajectories to define where the model is accurate to compute plans to the goal. In this example, the known dynamics model on the upper left $(\hat{f}(s,a,s'))$ reasons only about the containers but not about the plant. The problem is to define where the model is accurate enough to compute plans to the goal. The resulting learned model precondition is then used at test time to only perform actions in the model precondition.}
    \label{fig:activemdethumbnail}
\end{figure}

We illustrate the intuition of our problem setting and approach in Fig.~\ref{fig:activemdethumbnail}. Given a dynamics model that is accurate in only some combinations of states and actions, the objective is to estimate model deviation in state-action space to define the model precondition. During each iteration, the robot collects data and updates its model precondition based on that data, which is a form of active learning. Although we can draw from existing active learning techniques in other robotics applications~\cite{wang2021learning}, collecting a useful MDE dataset is challenging because it needs to contain a diverse set of trajectories to both identify the limits of the model and reliably solve planning problems. Furthermore, model error in earlier states of a selected trajectory can lead the robot to be unable to gather data from the later states due to the sequential nature of the problem.

To address these problems, we describe active learning techniques for efficiently selecting trajectories to learn the MDE. We propose a task-oriented approach for generating trajectory candidates, as well as multi-step acquisition functions that compute a single utility value from the sequence of transitions in a trajectory. Our approach enables sample-efficient learning of an MDE for solving tasks from a given distribution. We analyze how variations on the active learning algorithm affect the dataset, and subsequently the quality of the model preconditions at test time.

This paper makes the following contributions: (1) a novel problem formulation and approach for active learning of model preconditions defined by MDEs to improve the accuracy and robustness of plans for manipulation tasks with variable-accuracy models; (2) analysis of the effect of acquisition function choices on trajectory selection during training and the resulting test-time reliability of executed plans.

\section{Related Work}
This work focuses on planning with variable-accuracy models, where assuming globally high accuracy would lead to failures.
Existing techniques to mitigate the impact of model deviation on task performance such as adaptive control~\cite{narendra1997adaptive,fu2016one}, and reasoning about uncertainty with probabilistic models~\cite{pall2018contingent, kaelbling2013integrated, levine2014learning} are still susceptible to inaccuracy~\cite{guo2017calibration}. Furthermore, some models, including simulators and many analytical models, lack the capacity to represent uncertainty. Similarly, dynamics models learned from data~\cite{takano2022,nagabandi2020deep,hafner2019learning, wu2023daydreamer} can be inaccurate for a variety of reasons such as scenarios outside the training data distribution and limited model capacity for complex interactions. Despite these limitations, such models demonstrate practical utility in various planning tasks.

Our approach does not intend to replace learned or uncertain dynamics models, but rather to complement them to address persistent model inaccuracies. Other work has shown that estimating model deviation can be more data-efficient than learning a dynamics model with an equivalent amount of data, and lead to higher reliability~\cite{power2021keep, liu2005model, UnreliableMitrano2021}. \ed{2}{Despite some data-efficiency improvements, current approaches to estimating model deviation lack active learning capabilities, limiting their use to scenarios where the inherent randomness of the planning process and environment can sufficiently cover the data space ~\cite{mitrano2022data, UnreliableDale2019}. }

\ed{10}{Though we address a different problem, we use similar tools as broader active learning techniques used in other areas of robotics, such as using probabilistic models to select informative samples for dynamics models, skill preconditions, and policies ~\cite{abraham2019active,wang2021learning, eysenbach2021maximum}. } Active dynamics learning~\cite{capone2020localized, buisson2020actively} approaches sometimes address the additional challenge of sequential dependence of selecting informative points, but these works put strong assumptions on the form of the dynamics model, whereas we extend our scope to allow model preconditions over various types of models such as analytical models and simulators. Model preconditions are different but potentially more generalizable than skill preconditions since \ed{7}{multiple model-based behaviors can use the same model preconditions}. 

Real-world fluid manipulation particularly benefits from efficient exploration. Existing works tend to be conservative in action space by limiting pours to a small region, such as directly over a target container, which greatly limits the set of observable dynamics~\cite{correll2010indoor,7041448, schenck2017visual}. Furthermore, other approaches are largely constrained to scenarios with simple dynamics~\cite{kennedy2019autonomous, noda2007modeling, schenck2017visual, vaz2020model} where failure tends to be over-pouring or under-pouring. To our knowledge, this is the first experimental setting that uses a commonly-used 7 DOF manipulator to perform actions that can often spill water into the workspace. 

\section{Problem Statement}
\label{sec:problemstatement}

In this work, we actively learn model preconditions for planning with inaccurate dynamics models of the form $\hat{s} \gets \hat{f}(s,a)$. We do \emph{not} make additional assumptions on the implementation or source of the model (e.g. analytical model, simulator, learned model). A model precondition, denoted as $\mathrm{pre}(\hat{f})$, is a region where a planner may use a given $\hat{f}(s,a)$. 
 
The planning problems in our setting are defined by sampling a start state and goal function $g(s)$ that outputs whether or not $s$ is a goal state.
Goals are achieved by planning and executing a trajectory defined by actions $a_{1:T-1}$ and predicted states $\hat{s}_{1:T}$ such that $g(s_T)$ holds.
 We assume that $\hat{f}(s,a)$ is sufficient for solving the planning problems. At test time, the planner uses the learned model precondition to reject transitions where $(s,a) \notin \mathrm{pre}(\hat{f})$. 

The concrete form of model preconditions we use describes $(s,a,s')$ transitions where the deviation between predicted states and next states $d(\hat{s}, s')$ stay within a threshold tolerance, $\dmax$, that the system can tolerate or correct. $d(s_i,s_j)$ is a distance function, such as Euclidean distance, that outputs a scalar. The constraint can then be defined as $\mathrm{pre}(s,a) = \{s,a \,|\, d(s', \hat{f}(s,a)) < \dmax \}$. Since $d(s', \hat{f}(s,a))$ is impossible to compute without knowing $s'$, we instead estimate $d(s', \hat{f}(s,a))$ given $(s,a)$, denoted as $\hat{d}(s,a)$ to indicate that it is estimated for a state and action. 

The active learning problem in this work is to select a set of (variable-length) trajectories to form a dataset $\mathcal{D}$ of $(s,a,s')$ tuples on which an MDE is trained. 
Each trajectory is denoted by $\tau$, and we describe the set of candidate trajectories that different searches generate for the same problem as $\trajset$.
The agent may use the planner during training time and sample from the same \textit{distribution} of planning problems that will be seen at test time, but not the same problems.
We assume access to sufficiently accurate state estimation to compute meaningful deviations between all observed points on the trajectory. 

\section{Learning a Model-Deviation Estimator}
\label{sec:learnmde}

\newcommand{\labels}{d(\hat{f}(s,a), s')}
As defined in Section~\ref{sec:problemstatement}, an MDE predicts the deviation $d(\hat{s}, s')$ for a particular model $\hat{f}(s,a)$ between a predicted state $\hat{s}$ and the true next state $s'$. We denote the output of an MDE as $\hat{d}(s,a)$. \ed{1}{By directly predicting $d(\hat{s}, s')$, the MDE is agnostic to the source of model deviation.} The MDE in this paper is a Gaussian Process (GP) model with a Mat\'ern kernel and heteroscedastic noise model, specifying a Gaussian distribution for the deviation with mean $\mu(\hat{d}(s,a))$ and standard deviation $\sigma(\hat{d}(s,a))$. \ed{1}{A heteroscedastic noise model enables input-dependent noise, which is important to capture when sources of model deviation differ}. 

Data collected for the MDE is in the form of $(s, a, s')$ tuples from executing action $a$ in the target environment (i.e., the real world) from state $s$, and then observing $s'$. The input to the MDE is $(s,a)$ and the label is $d(\hat{f}(s,a), s')$.
\begin{figure*}[t!]
    \centering    \includegraphics[width=1.0\textwidth]
    {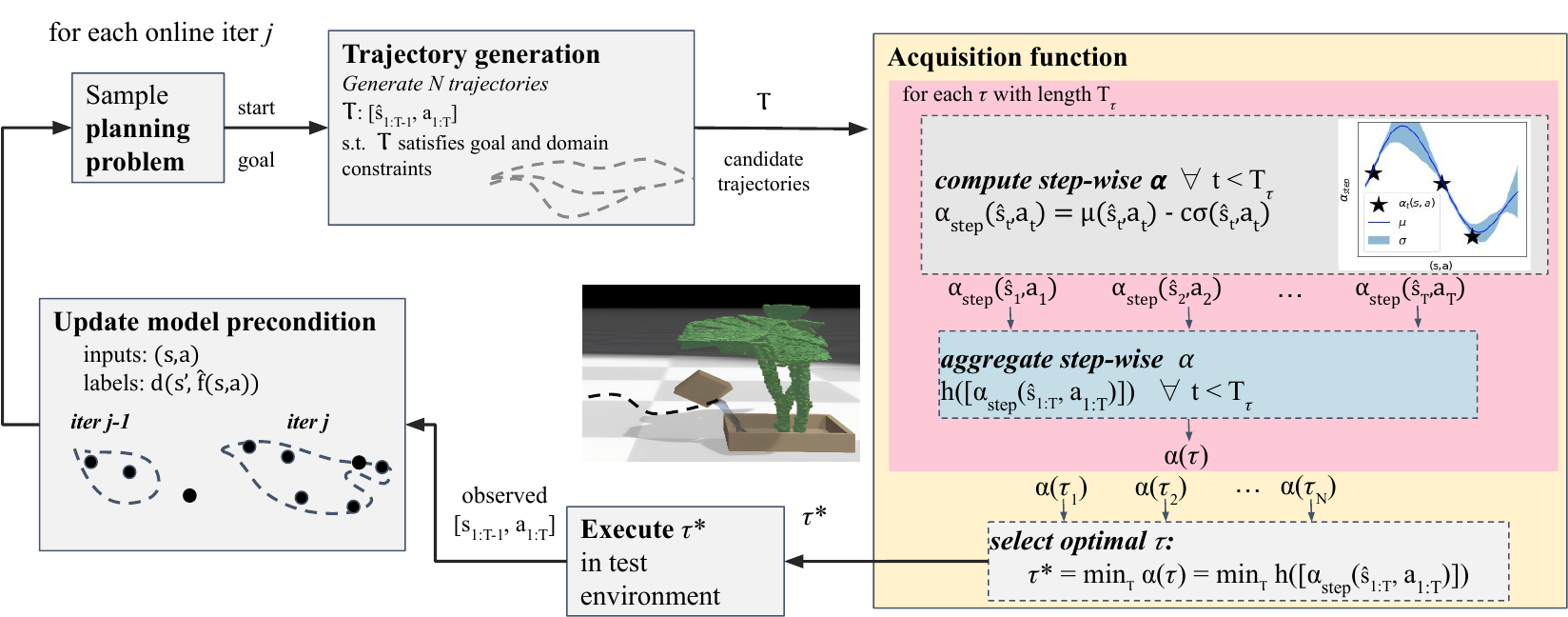}
    \caption{Overview of our method: Each iteration $j$ starts with sampling a planning problem and generating candidate trajectories that satisfy domain constraints and reach the goal. 
    We outline the acquisition function computation for \emph{each} trajectory in the pink box, including the step-wise acquisition function values, $\alpha_{\mathrm{step}}(s_t,a_t)$ for each state-action pair in the trajectory. These values are then aggregated by a function $h$ to yield the trajectory's utility: $\alpha(\tau)$
. The final step is selecting and executing $\tau*$, in the test environment to collect the ground truth $[s_{[1:\trajlen]}, a_{[1:\trajlen-1]}]$. The MDE is updated every $M$ trajectories. }
\vspace{-0.1cm}
    \label{fig:overview}
\end{figure*}

A learned MDE then defines a model's precondition as: $\mathrm{pre}(\hat{f}) = \{(s,a) | P(\hat{d}(s,a) > \dmax) < \delta\}$ for some small probability $\delta$.
The constraint can be written as $\mu(s,a) + \beta \sigma(s,a) < \dmax$, where higher $\beta$ lowers the risk tolerance.


\section{Active Learning}
\label{sec:activelearning}

The algorithm for actively learning MDEs is illustrated in Fig.~\ref{fig:overview}. First, the agent samples a planning problem and then uses a motion planner to generate candidate trajectories, $\trajset$. \ed{1}{The search adds transitions to the tree if they satisfy planning constraints such as joint limits, a collision check, and the model precondition as defined in Section~\ref{sec:learnmde}. To encourage exploration, a zero-mean prior is used for the MDE during the learning phase. We use a rapidly exploring random tree (RRT) planner~\cite{lavalle1998rapidly} to encourage a diverse set of solutions in $\trajset$.  
Then, the robot executes the trajectory that minimizes an acquisition function $\alpha(\tau)$, which is a heuristic for the utility of $\tau$ to the MDE.} \ed{1}{After a batch of $M$ executed trajectories for $M$ problems, the robot adds the observed $(s,a,s')$ tuples to $\mathcal{D}$ and updates the MDE using the training method described in Section~\ref{sec:learnmde}. Ideally, the model precondition region between the start and goal states expands with more data (Fig.~\ref{fig:activemdethumbnail}). }

\ed{3}{At test time, the robot generates a trajectory to the goal using the same planner, MDE, and constraints as during training, but with a more conservative model precondition. The $\beta$ parameter as described in Section~\ref{sec:problemstatement} sets the tolerance for deviations outside the model precondition; for example, $\beta=2$ specifies a 98\% confidence interval.}

\subsection{Acquisition Function}

\newcommand{\alphastep}{\alpha_{\mathrm{step}}}

Now, we explain how we define the acquisition function $\alpha(\tau)$, which guides our selection of the trajectory to execute in each iteration: $\tau^* \gets \mathrm{argmin}_{\tau \in \trajset} ; \alpha(\tau)$. The procedure is illustrated in the rightmost box of Fig.~\ref{fig:overview}.

\textbf{Step-wise utilities:}  First, we compute utilities for each step in the trajectory, shown in the dotted box. 
Trajectories can vary in length, denoted as $\trajlen$, and are comprised of states and actions: $s_{1:\trajlen}, a_{1:\trajlen-1}$.
We denote the utility for each step $t$ as $\alphastep(s_t, a_t)$ and define it using a form inspired by Lower Confidence Bound: $\mu(x) - c\sigma(x)$ where $c$ controls exploration.

\newcommand{\reductionseq}{[\alphastep(s_t, a_t)\; \forall \; t \;< \;\trajlen]}
\textbf{Aggregating individual step-wise utilities for a trajectory:}  As shown in the blue dotted box (Fig.~\ref{fig:overview}), we next define a function $h$ that aggregates single-step utilities in trajectories of different lengths, $\alpha_{step}(s_{1:\trajlen}, a_{1:\trajlen})$, to a single trajectory utility, $\alpha(\tau)$. 

The general form of a trajectory-based acquisition function using the lower confidence bound-based $\alphastep$ is thus:
\begin{align}
  \alpha(\tau) \gets h([ \mu(\hat{d}(s_t,a_t)) - c\sigma(\hat{d}(s_t,a_t))] \; \forall t < \trajlen) 
\end{align}

Since later transitions may not be reached when the trajectory is planned using an inaccurate dynamics model, we introduce $h_{\mathrm{max}}$ and $h_{\mathrm{sum}}$.  $h_{\mathrm{max}} = \mathrm{max}_{t < \trajlen} \gamma^ t \alphastep(s_t,a_t)$ and $h_{\mathrm{sum}} = \sum_{t < \trajlen} \gamma^t \alphastep(s_t,a_t)$. Multiplying each $\alphastep(s_t,a_t)$ by $\gamma^t$ approximates the idea that nearer steps are more useful in the trajectory.

\subsection{Candidate trajectory generation}
\label{sec:betasked}
To set the risk tolerance during training, we propose a schedule for the MDE that gradually reduces risk tolerance as the robot accumulates more data. Since $\delta$ is determined by $\mu(s,a) + \beta \sigma(s,a)$, $\delta$ can be set by modifying $\beta$ using the inverse CDF of a Gaussian distribution: $\beta = \Phi^{-1}(1 - \delta)$.  At iteration $j$ of $J$ total iterations, $\beta_j \gets \frac{2k_1}{1+\mathrm{exp}\bigl(-k_2(j-\frac{J}{2})\bigr)}-k_1$ results in a transition from $-k_1$ to $k_1$ where a lower $k_2$ causes a smoother transition. 

\newcommand{\plotwidth}{0.24\textwidth}
\newcommand{\ra}{\textit{Random}\xspace}
\newcommand{\rp}{\textit{Goal-Conditioned}\xspace}
\newcommand{\ac}{\textit{Active Cautious}}
\newcommand{\amo}{\textit{Active Mean Only}}
\newcommand{\ab}{\textit{Active Goal-Conditioned}\xspace}
\newcommand{\lowsuccess}{\xspace{\small\texttt{below\_success}}\xspace}
\newcommand{\lowfail}{\xspace{\small\texttt{below\_spill}}\xspace}
\newcommand{\highsuccess}{\xspace{\small\texttt{above\_success}}\xspace}
\newcommand{\highfail}{\xspace{\small\texttt{above\_spill}}\xspace}
\newcommand{\watersimenvname}{\xspace{\small\texttt{Water Plant(sim)}}\xspace}
\newcommand{\waterrealenvname}{{\small\texttt{Water Plant(real)}}\xspace}
\newcommand{\gridworldname}{\xspace{\small\texttt{Icy GridWorld}}\xspace}

\newcommand{\onlinelearningresultssubfigs}[1]{
        \begin{subfigure}{\plotwidth}
            \includegraphics[width=\textwidth]{imgs/activelearningplots/#1/plan_found.png} 
        \end{subfigure}
        \begin{subfigure}{\plotwidth}
            \includegraphics[width=\textwidth]{imgs/activelearningplots/#1/success.png} 
        \end{subfigure}
        \begin{subfigure}{\plotwidth}
            \includegraphics[width=\textwidth]{imgs/activelearningplots/#1/tpr.png} 
        \end{subfigure}
        \begin{subfigure}{\plotwidth}
            \includegraphics[width=\textwidth]{imgs/activelearningplots/#1/tnr.png} 
        \end{subfigure}
}

\newcommand{\onlinelearningresultsfig}[3]{
    \begin{figure}
        \centering
        \onlinelearningresultssubfigs{#1}
        \caption{#2}
        \label{#3}
    \end{figure} 
}

\newcommand{\twodomainonlinelearningresultsfig}[6]{
    \begin{figure*}[h!]
        \centering
        \begin{subfigure}{\textwidth}
            \onlinelearningresultssubfigs{#1}
            \caption{#2}
        \end{subfigure}
        \begin{subfigure}{\textwidth}
            \onlinelearningresultssubfigs{#3}
            \caption{#4}
        \end{subfigure}
        \caption{#5}
        \label{#6}
    \end{figure*} 
}
\newcommand{\threedomainonlinelearningresultsfig}[8]{
    \begin{figure*}[ht!]
        \centering
        \begin{subfigure}{\textwidth}
            \onlinelearningresultssubfigs{#1}
            \caption{#2}
        \end{subfigure}
        \begin{subfigure}{\textwidth}
            \onlinelearningresultssubfigs{#3}
            \caption{#4}
        \end{subfigure}
        \begin{subfigure}{\textwidth}
            \onlinelearningresultssubfigs{#5}
            \caption{#6}
        \end{subfigure}
        \caption{#7}
        \label{#8}
    \end{figure*} 
}

\section{Experimental Setup}
\label{sec:evalmethodology}

\begin{figure}[h]
    \centering
    \includegraphics[width=1\textwidth]{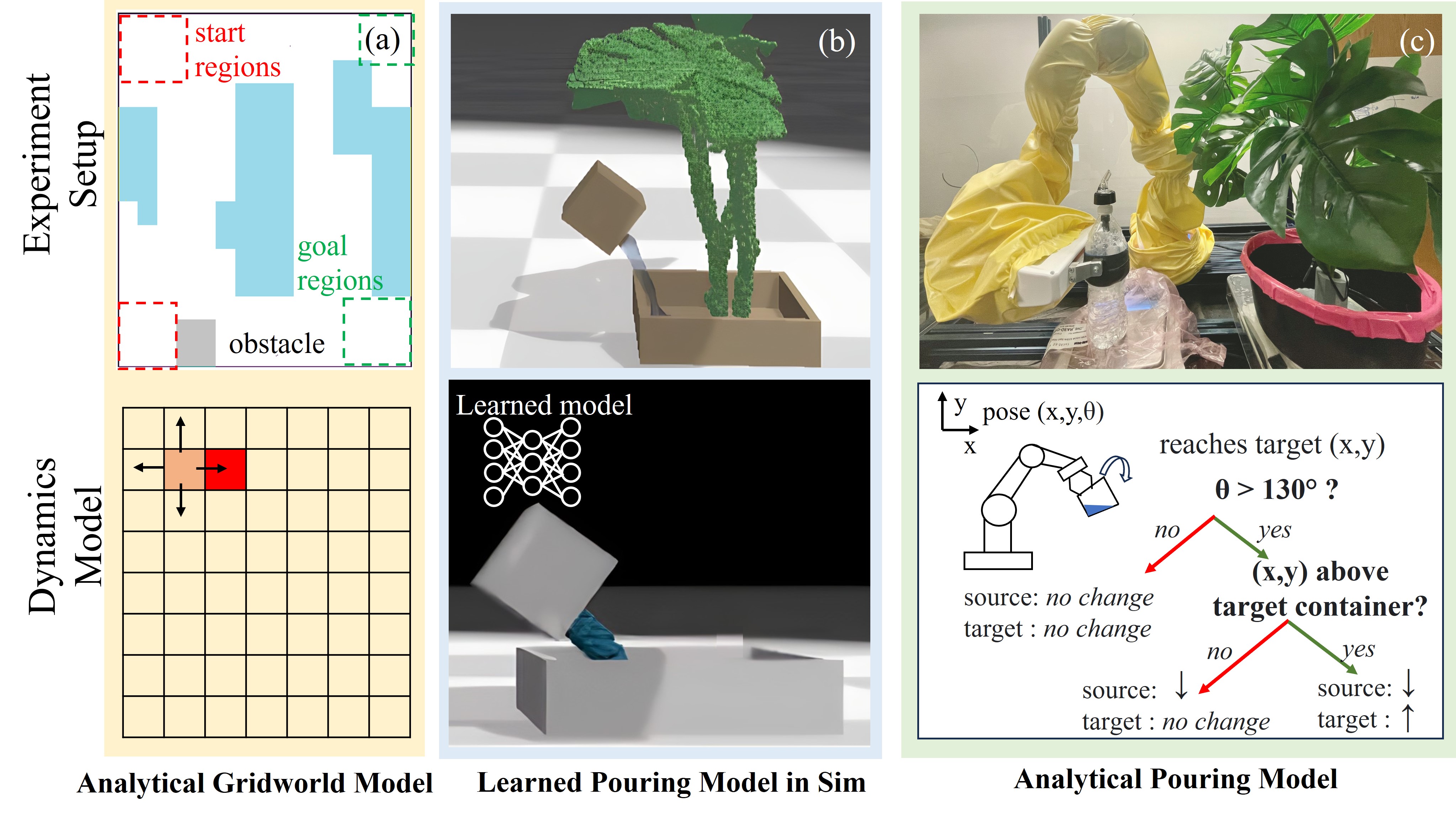} 
    \caption{ Scenarios and their corresponding dynamics models. (a) Slippery grid world where movement may result in slipping backwards over ice (blue) or not moving (grey). The analytical dynamics model assumes unimpeded movement within grid bounds. (b) Simulated plant watering using a learned dynamics model trained on a scenario without a plant. (c) Real-world plant watering with a rule-based analytical dynamics model.} 
    \label{fig:experiments}
    \vspace{-10pt}
\end{figure}

The first scenario, \gridworldname(Fig.~\ref{fig:experiments}a), is a gridworld also used in~\cite{vemula2020planning} where the robot can move in four cardinal directions, but if it moves left or right over an icy state, it slips by moving two cells backwards. The robot cannot move through the obstacle. The dynamics model assumes the robot moves to the intended location. The start and goal state are selected randomly for each planning problem. 

\newcommand{\trajTypeWidth}{0.32\textwidth}
\newcommand{\trajTypeImgWidth}{0.9\textwidth}
\newcommand{\trajTypeHeight}{2cm}
\begin{figure}[h]
    \centering
       \begin{subfigure}[b]{\trajTypeWidth}
       \centering
      \includegraphics[width=\trajTypeImgWidth]{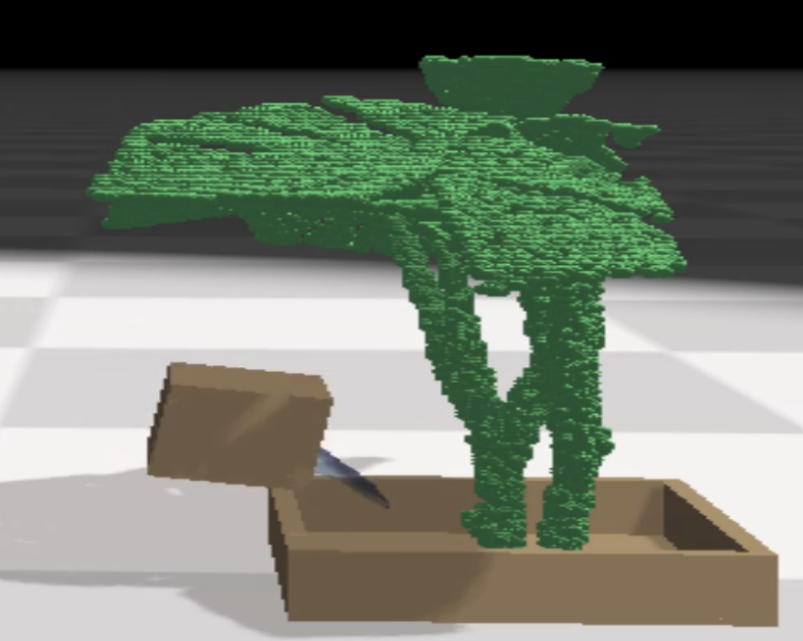} 
      \label{subfig:lowgood}
     \end{subfigure}%
     \hfill
    \begin{subfigure}[b]{\trajTypeWidth}
       \centering
      \includegraphics[width=\trajTypeImgWidth]{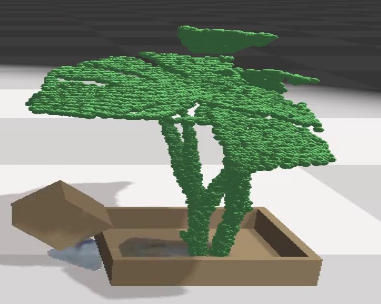} 
     \end{subfigure}%
     \hfill
    \begin{subfigure}[b]{\trajTypeWidth}
       \centering
      \includegraphics[width=\trajTypeImgWidth]{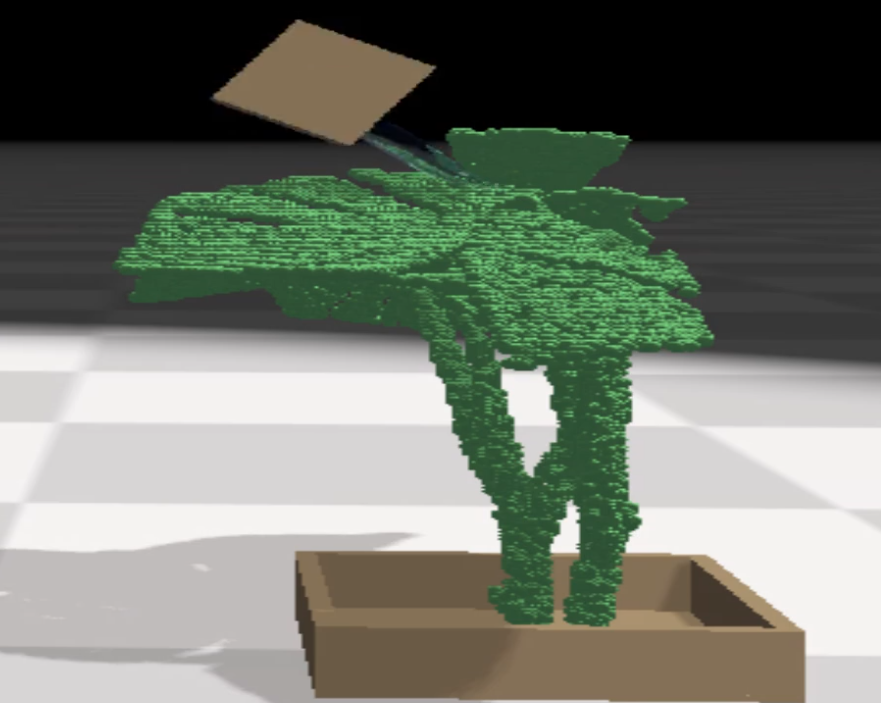} 
     \end{subfigure}
       \begin{subfigure}[b]{\trajTypeWidth}
      \includegraphics[width=\linewidth]{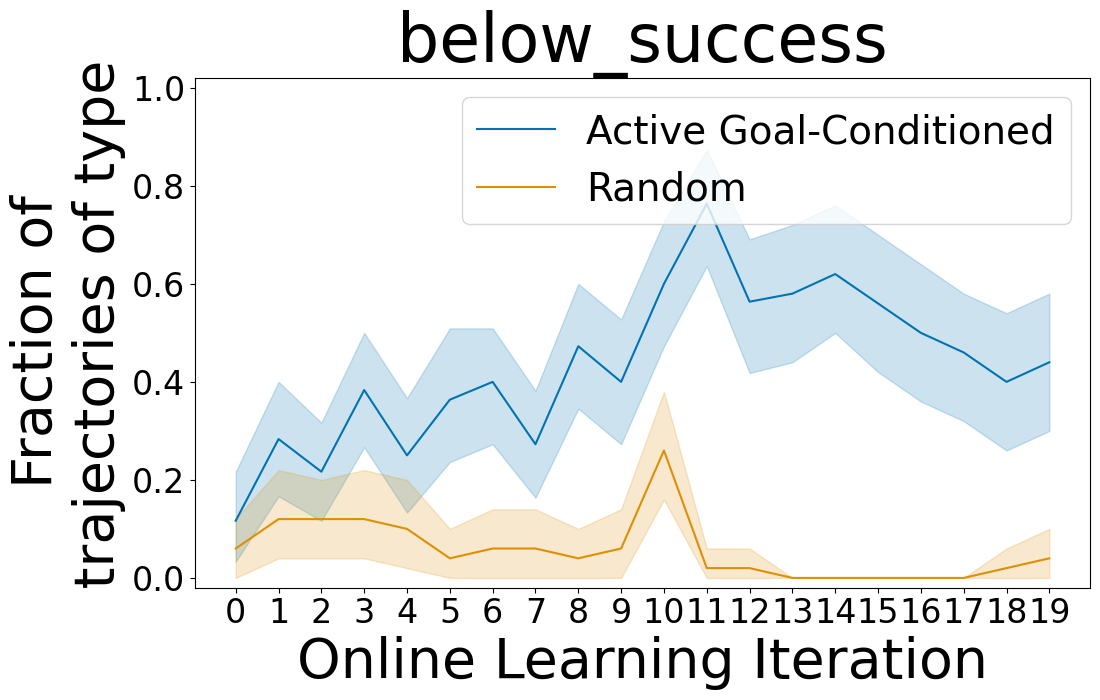} 
      \caption{\emph{below} leaves success}
      \label{subfig:lowgood}
     \end{subfigure}%
     \hfill
       \begin{subfigure}[b]{\trajTypeWidth}
      \includegraphics[width=\linewidth]{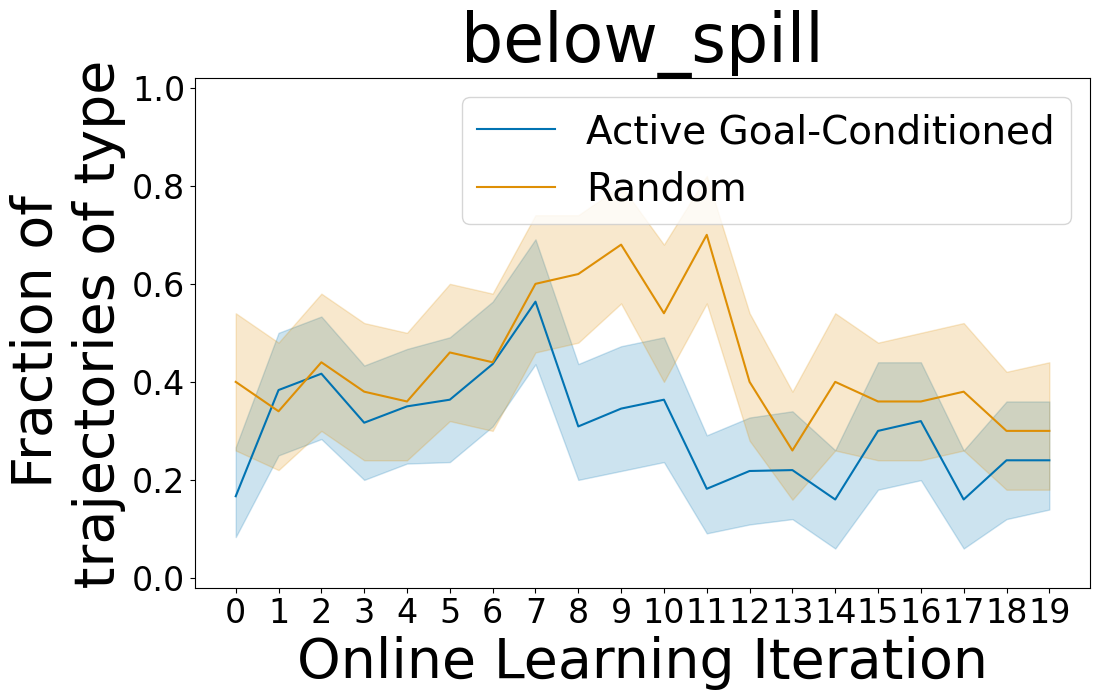} 
      \caption{\emph{below} leaves and spills}
      \label{subfig:lowgood}
     \end{subfigure}%
     \hfill
    \begin{subfigure}[b]{\trajTypeWidth}
      \includegraphics[width=\linewidth]{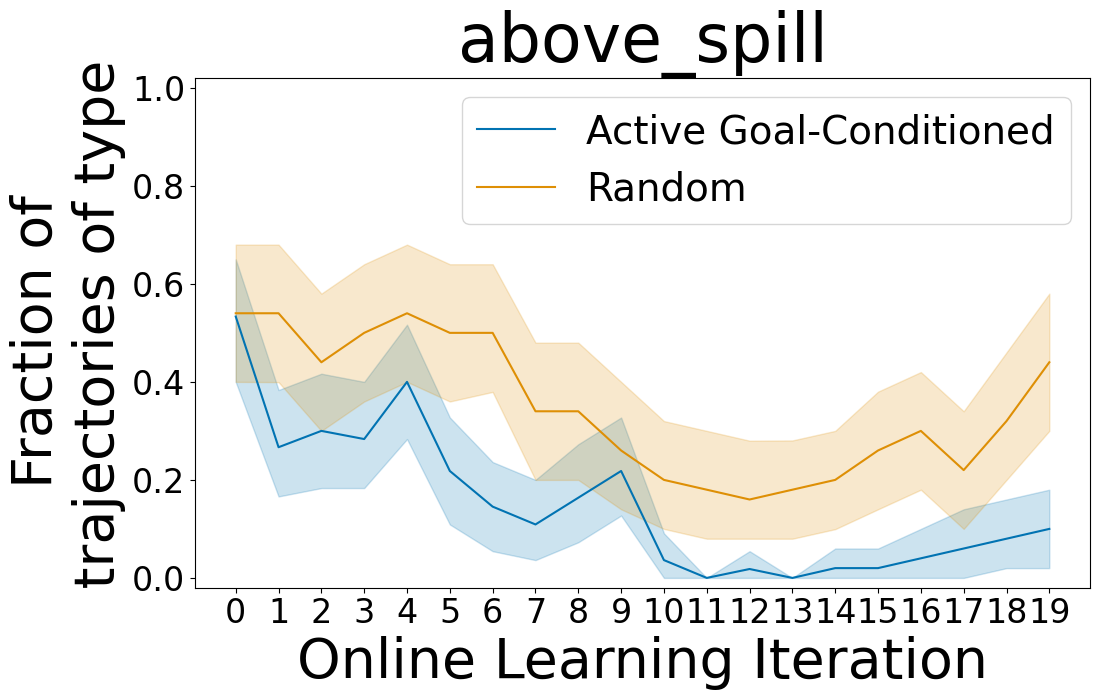} 
      \caption{\emph{above} leaves and spills}
     \end{subfigure}%
    \caption{Ratio of trajectory types executed during training (examples shown above) for our method and the \ra ablation over training iterations.}
    \label{fig:trajtypes}
\end{figure}
The second scenario, \watersimenvname (Fig.~\ref{fig:experiments}b), is a plant-watering domain where the goal is to pour a specified amount of water from a source container into a target container without spilling more than 2\%. The state space is the poses of both containers and their liquid volumes. The actions are specified as a target pose for the source container. The actions must satisfy constraints unrelated to the dynamics model including being collision-free according to an approximate collision checker, and either translating or rotating in one motion, but not both. The model is a neural network dynamics model trained in a simpler environment (shown in the bottom half of Fig.~\ref{fig:experiments}) with no plant and a wider source container. This scenario tests the algorithm's ability to learn model preconditions caused by multiple sources of model error such as geometry mismatch, obstruction by leaves, and unexpected collisions. The start state is a random pose left of the target container. 

Lastly, the third scenario, \waterrealenvname (Fig.~\ref{fig:experiments}c), is a real-world variation of the previous plant-watering domain.  A measured pourer dispenses 15 mL of water when tilted above 130 degrees. The action space and state space representations are consistent with the simulated scenario, but we restrict the MDE input to only the action to reduce dimensionality. 
\ed{11}{The branches can move, but the base stays fixed relative to the container. Since the plant state is only measured by the container pose, variations that affect the dynamics cause noise.}
This scenario demonstrates that reliable performance can be reached in a small number of trajectories (less than a dozen) in the real world where there is considerably more noise and variation. The analytical model we use assumes that 15 mL is dispensed for rotational actions above 130 degrees and that the water enters the target container if poured above the area of the container. The start state is fixed and the goal is for 15 mL to be in the target container without spilling more than 5 mL. 

\textbf{Evaluation methodology: } On simulated domains, we evaluate each variation using 10 seeds for 20 learning iterations. The simulated domains use five training trajectories per iteration, and the real domain uses two. \ed{4}{Setup, planning and execution for each training trajectory takes 15 s in \gridworldname, 1.5 min in \watersimenvname, and 3 m in \waterrealenvname.}  At test time, we evaluate the model preconditions for each iteration by using the model preconditions where $\hat{d}(s,a) < \dmax$ for each transition with 98\% confidence. In the simulated scenarios, we sample 20 planning problems per iteration, and in the real-world scenario, we sample 5 per iteration.  In \waterrealenvname, we only evaluate the effect of using active learning and goal-conditioned candidate trajectories. $\dmax = 0.1$ for all scenarios and represents the sum of all position distances over all objects. We use a risk-tolerance schedule (Section ~\ref{sec:betasked}) with $k_1=2$ and $k_2=\frac{1}{2}$  In both watering scenarios, the average volume deviation of both containers is added to the position error. For consistency, we scale the volume units between simulation and the real-world such that one unit is poured out. 

The metrics that we evaluate are as follows. First, we evaluate model precondition accuracy on a cross-validation dataset of trajectories from the other seeds. We measure both the true negative rate (\tnr) and the true positive rate (\tpr) of whether an individual $(s,a,s')$ data point is in the model precondition. A higher rate is better for both metrics, but the TNR is more important than the TPR because the model precondition only needs to cover enough state-action space to compute a plan. Second, we test whether the model precondition is sufficient to reliably compute plans by measuring the success rate of the planner within a fixed timeout of 5000 extensions. Finally, we evaluate the end-to-end success rate in achieving the goal, which measures the effect of using the estimated model preconditions over the entire trajectory.

\section{Results}
\newcommand{\modelprecondwidth}{0.36\textwidth} 
\newcommand{\itersubfig}[1]{
    \begin{subfigure}[t]{0.1\textwidth}  
    \includegraphics[width=\textwidth]{imgs/iteration_#1_image.pdf}%
    \end{subfigure}%
} %
\begin{figure}[h!]
    \itersubfig{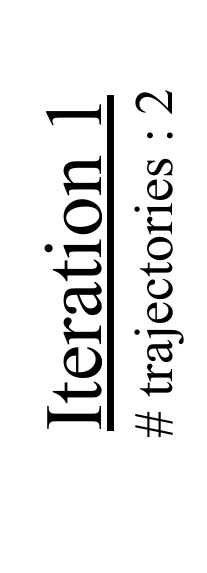}
    \begin{subfigure}[t]{\modelprecondwidth}  
      \includegraphics[width=\linewidth]{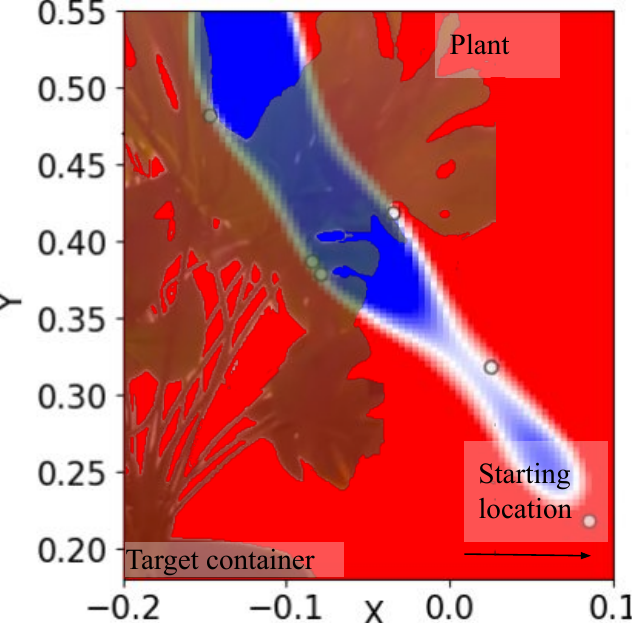} %
      \caption{Translation only}
     \end{subfigure}%
     \hfill
    \begin{subfigure}[t]{\modelprecondwidth} 
      \includegraphics[width=\linewidth]{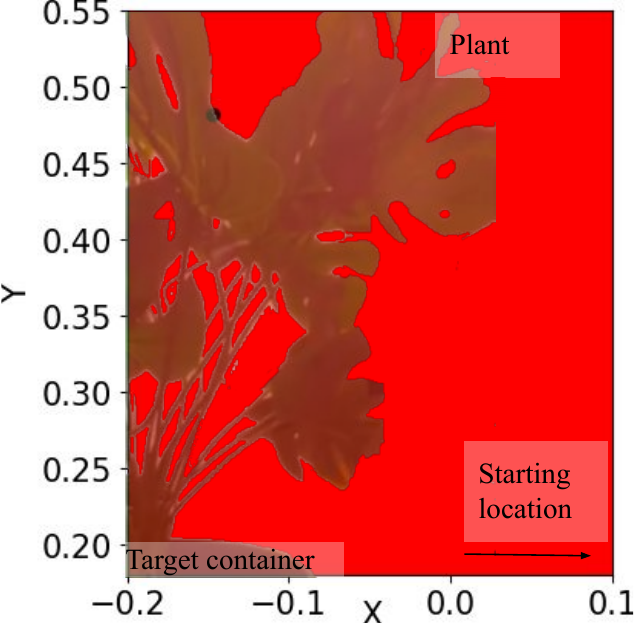} %
      \caption{Rotation only}
     \end{subfigure}%
    \begin{subfigure}[b]{0.13\textwidth} 
    \caption*{\underline{\textit{Key}}}
      \includegraphics[width=\linewidth]{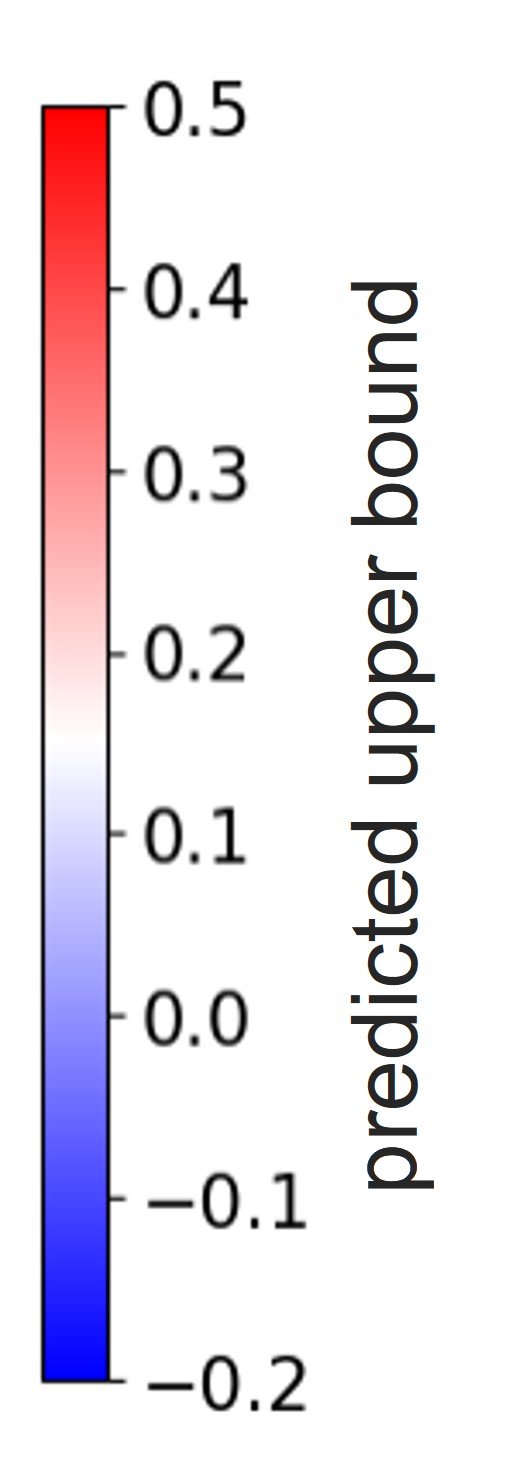} %
     \end{subfigure}
    \itersubfig{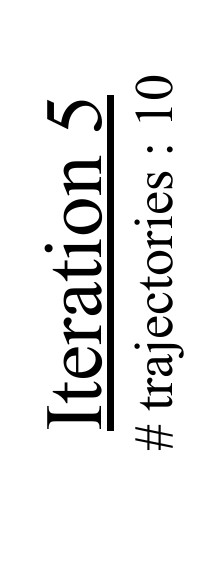}
    \begin{subfigure}[t]{\modelprecondwidth}
      \includegraphics[width=\linewidth]{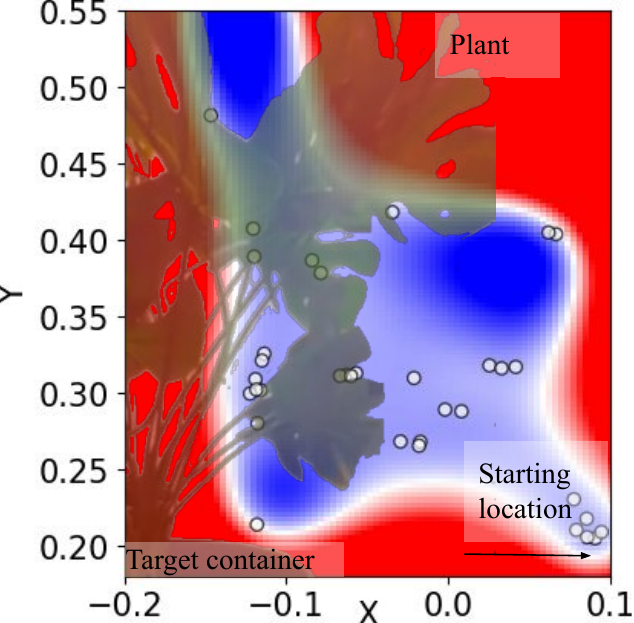} %
      \caption{Translation only}
     \end{subfigure}%
     \hfill
    \begin{subfigure}[t]{\modelprecondwidth}
      \includegraphics[width=\linewidth]{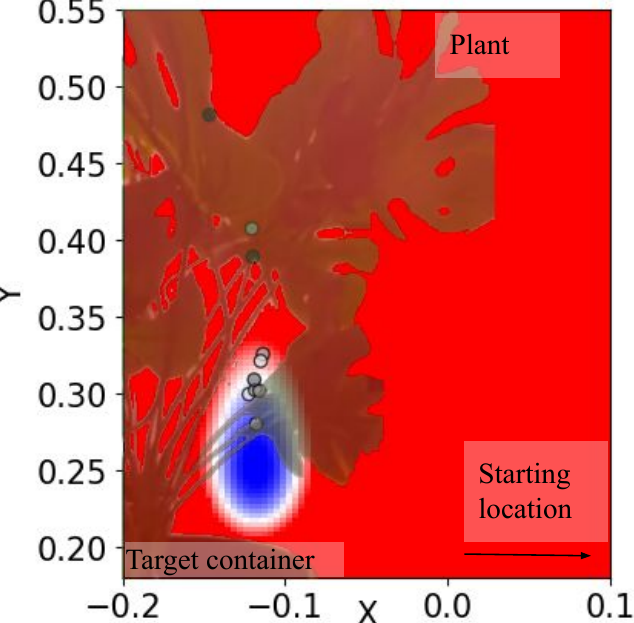} %
      \caption{Rotation only }
     \end{subfigure}%
     \hspace{0.01cm}
    \begin{subfigure}[t]{0.13\textwidth} 
      \includegraphics[width=\linewidth]{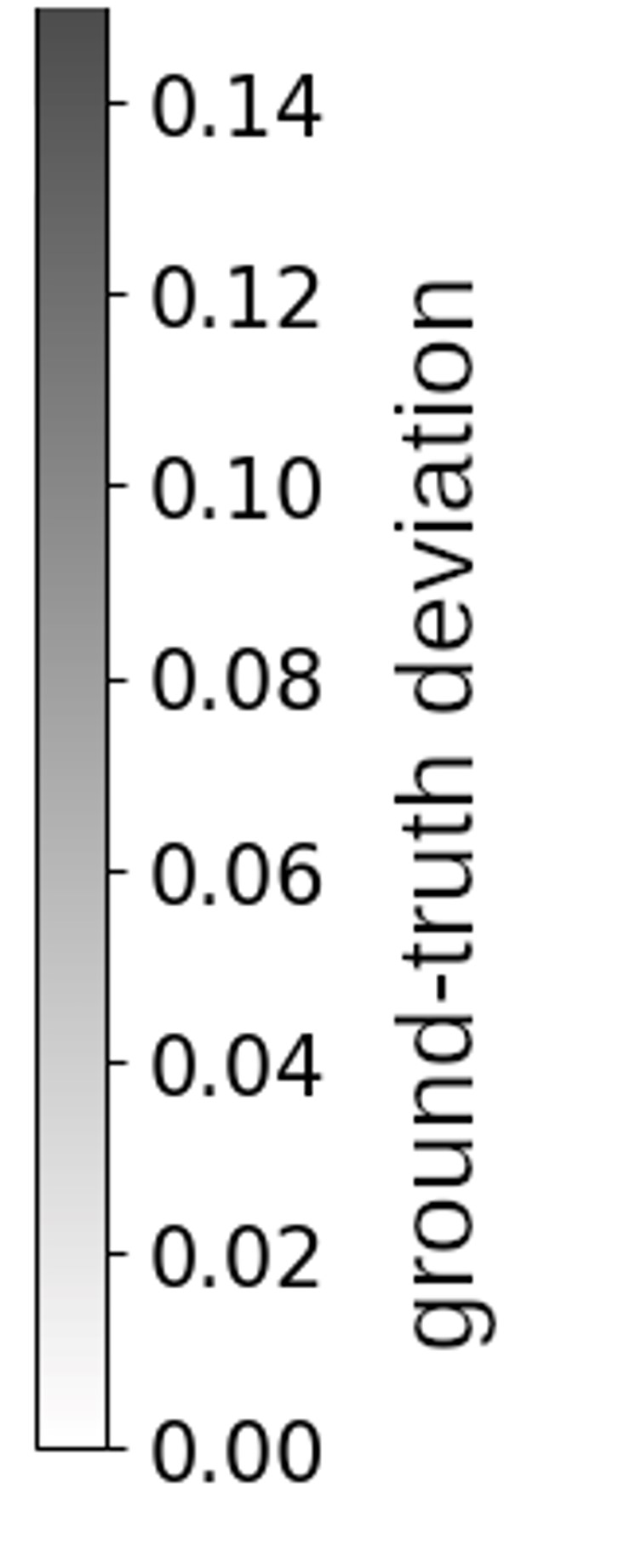} %
     \end{subfigure}
    \caption{MDE with plant overlay over training iterations for translation-only actions and rotation-only actions. Color scales indicate ground-truth deviation (right, top) and upper bound of the predicted deviation $\mu(\hat{d}(s,a)) + \beta \hat{d}(\sigma(s,a))$ for $\beta = 2$. $\dmax=0.1$, so the blue region indicates the model precondition.}
    \label{fig:modelpreconds}
\end{figure}
\threedomainonlinelearningresultsfig{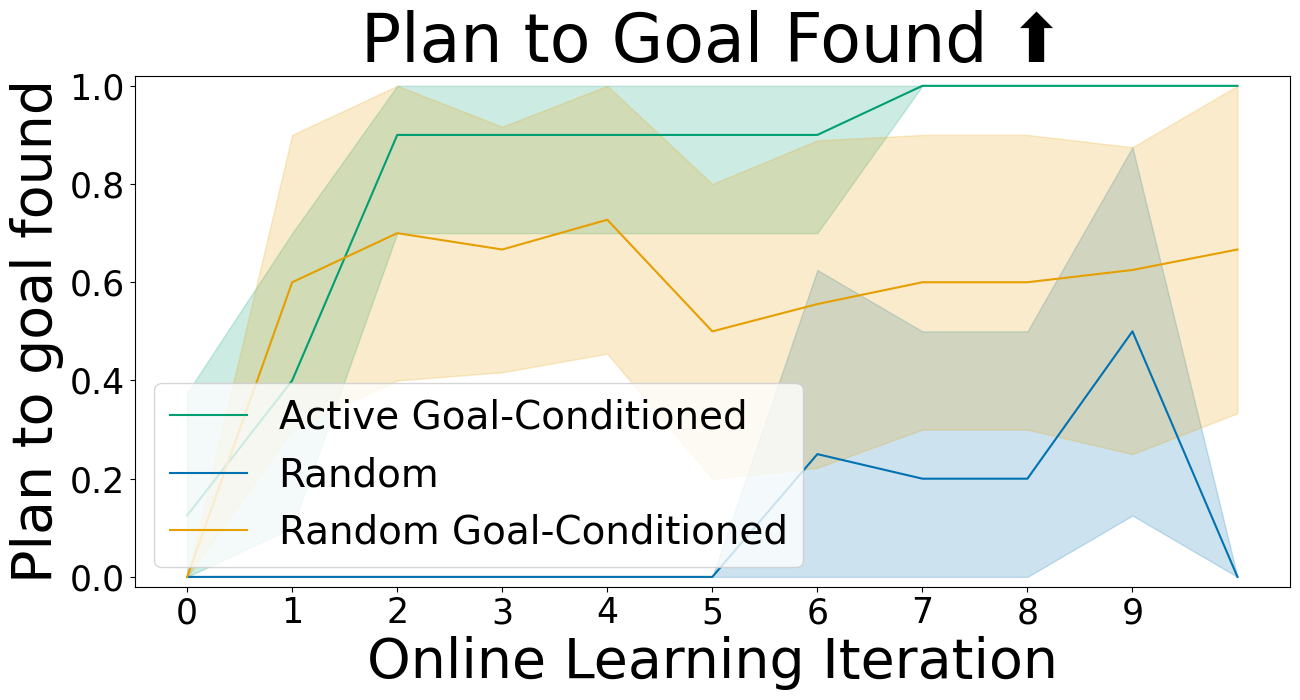}{\waterrealenvname}{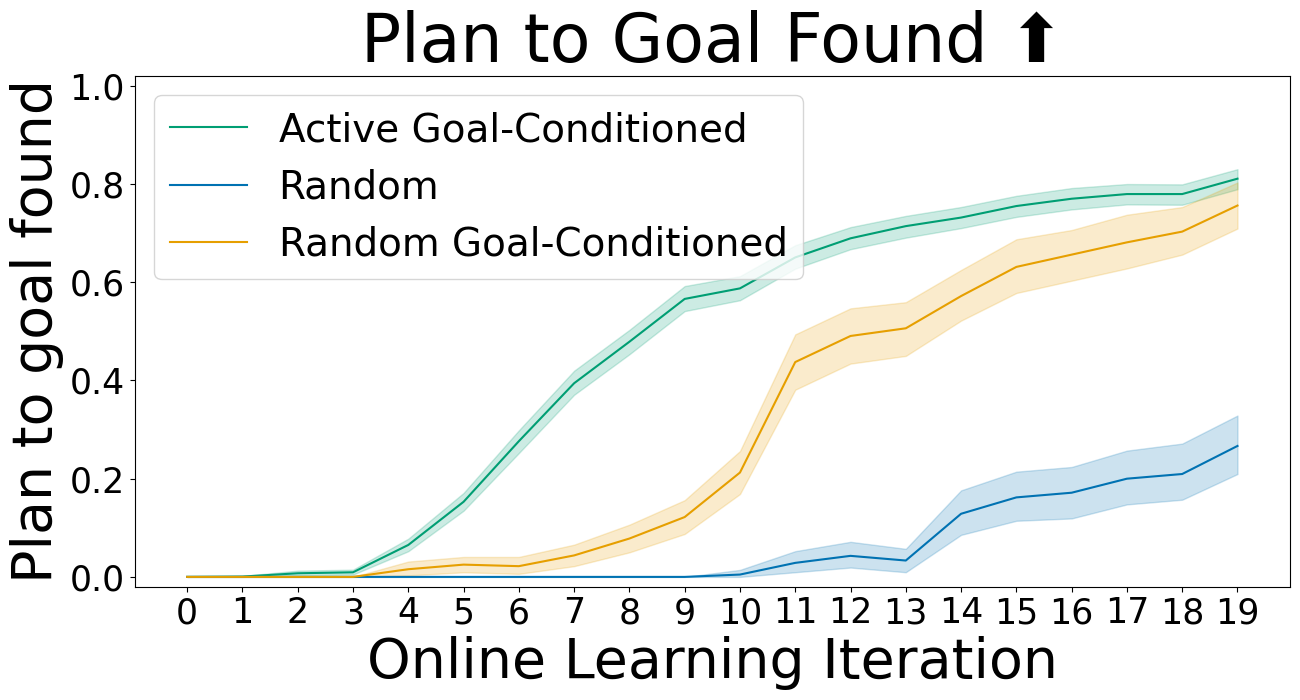}{\watersimenvname}{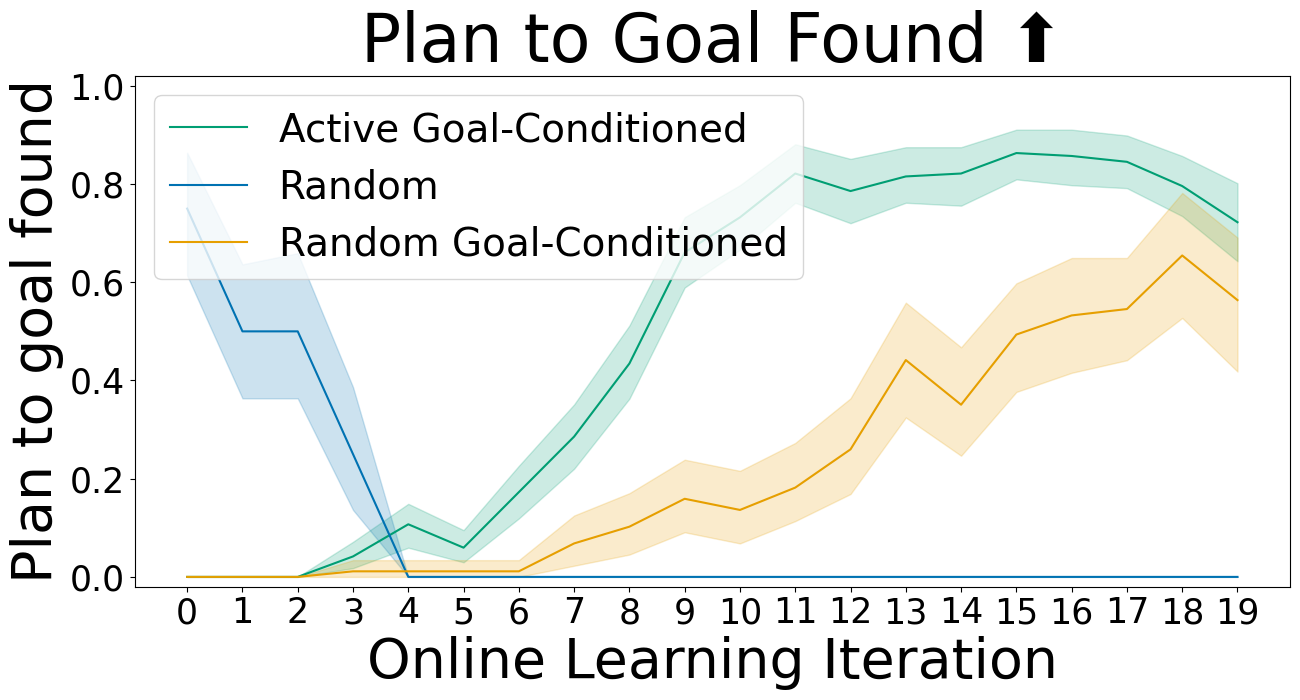}{\gridworldname}{Success rate in computing a plan, success rate in reaching the goal, and model precondition accuracy over training iterations for three scenarios. These plots measure the effect of whether trajectories are selected actively, goal conditioned, or not goal conditioned on performance.}{fig:goalconditioning}
We first show a qualitative analysis of data collected during the active learning. 
We then quantitatively evaluate the effects of different algorithmic variations using the metrics we previously described in Section~\ref{sec:evalmethodology}.
Two baselines using an MDE but different methods to generate $\trajset$ measure the impact of algorithmic choices. \ra selects random actions, effectively removing the goal-conditioning. \rp eliminates the acquisition function by selecting a random trajectory to the goal.
\twodomainonlinelearningresultsfig{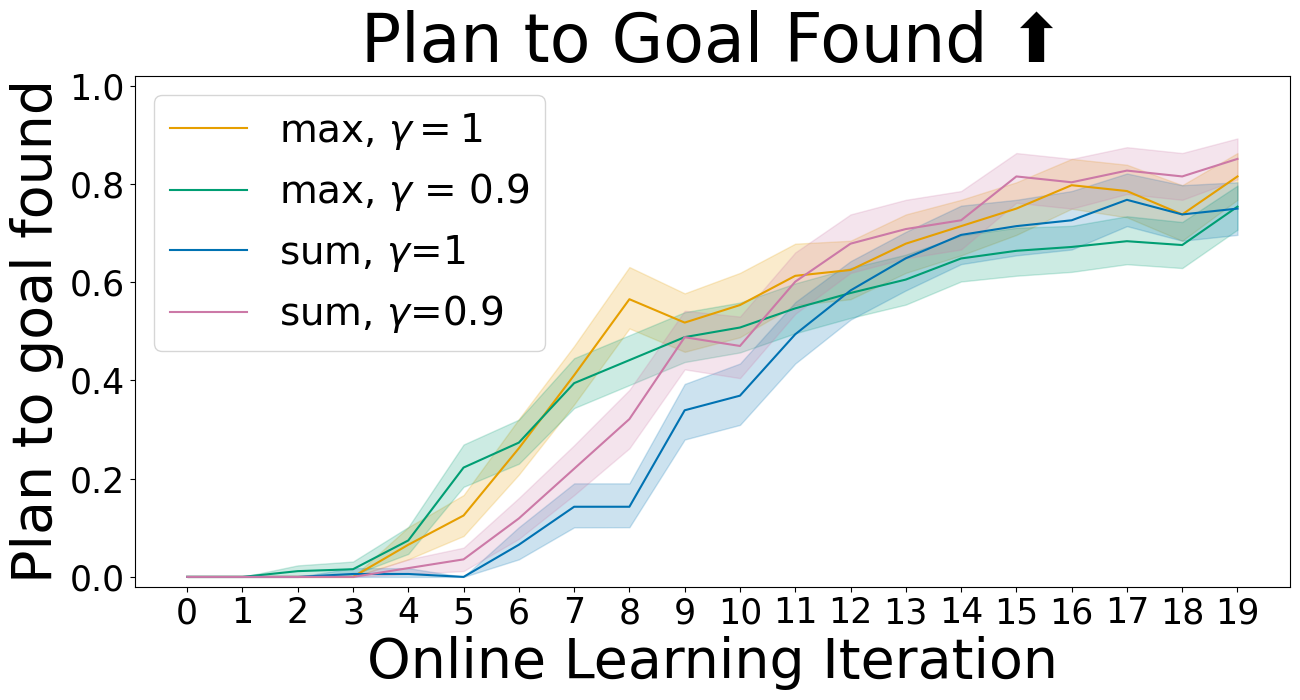}{\watersimenvname}{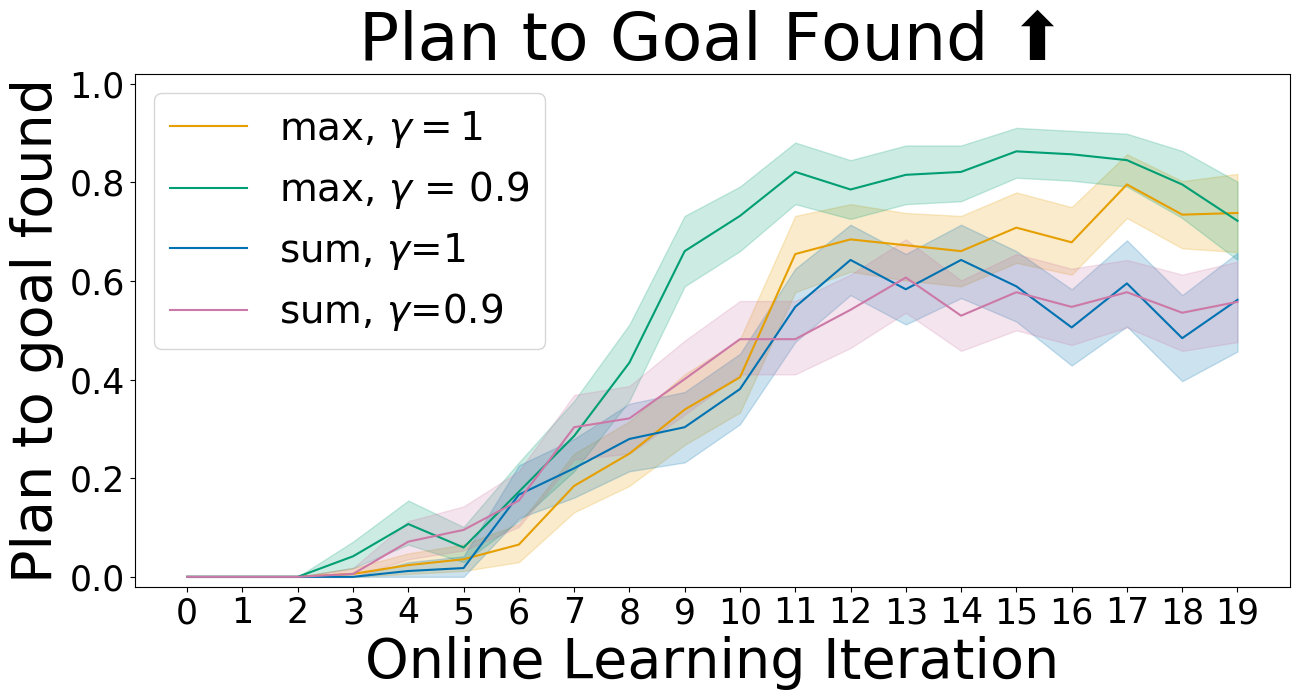}{\gridworldname}{
Success rate in computing a plan, success rate in reaching the goal, and model precondition accuracy over training iterations for three scenarios. These plots measure the effect of the aggregation function on performance.}{fig:reduction}

\textbf{Online learning analysis:} Here, we analyze the qualities of the MDE dataset when using our active learning approach described in Section~\ref{sec:activelearning} (\ab) compared to selecting trajectories that do not reach the goal, but use the same samplers from the planner, which we call \ra. The first row of Fig.~\ref{fig:trajtypes} shows labels of trajectories selected during training to analyze the makeup of the MDE dataset. \lowsuccess is the label for pours below the leaves where all the water reaches the target, which is what would ideally be well represented in the dataset. Since data where the model is inaccurate is also necessary to quantify the limitations of the model, we also track the portion of the dataset that is \lowfail. The final trajectory type we track is \highfail, which is the least useful because the water-container-leaf dynamics are typically high-deviation and unnecessary to complete tasks.

Fig.~\ref{fig:trajtypes} shows a faster increase in desirable \lowsuccess pours from \ab in iterations 0 to 9.
We note that there is an increase in successful pours around iterations 10-12, which then decreases. As we show in the included video, this effect can be attributed to responses in both aleatoric and epistemic uncertainty, which affects the exploration bonus: $c\sigma(\hat{d}(s,a))$.

\textbf{Model preconditions in \waterrealenvname:} Here, we visually analyze how regions of model preconditions change in the real-world pouring task. These regions are low dimensional so it can be visualized in 2D by splitting $\theta_d$ into two-cases: one for rotation-only actions, and one for translation-only actions. We show the model precondition by plotting $\hat{d}(s,a)$ over a region of $\mathcal{A}$ (Fig.~\ref{fig:modelpreconds}). 
Aligned with our intuition of how  model preconditions should evolve, the area of the model precondition expands as more data is collected. The boundary becomes more precise with more data points (Fig.~\ref{fig:modelpreconds}). By iteration 5, the model precondition for rotational actions is in a region above the target container but below the leaves. Note that despite a low-deviation point at iteration 5 at (-0.12, 0.39), that area is not in the model precondition because there is a nearby point that is high-deviation. 

%

\subsection{Candidate Trajectory Set Generation}

Here, we evaluate how the method to generate $\trajset$ affects performance metrics.
Fig.~\ref{fig:goalconditioning}, shows higher data efficiency when using \ab  or \rp in all scenarios. In \watersimenvname, we also see significantly higher data efficiency when using active learning in addition to conditioning on trajectories that reach the goal, indicated by higher performance after fewer iterations when using \ab  over using \rp, which does not use an acquisition function. 
In \waterrealenvname, we see a clear improvement when using goal-conditioning, and a modest improvement in success when using the acquisition function in later iterations. The improvement of \ab over \rp in finding plans to the goals is matched by a significant increase in the \tpr, which is more significant in the simulated environments than in \waterrealenvname. \ab shows an improvement in success rate reflected by a higher \tnr.
Earlier iterations in \gridworldname using \ra have overly broad model preconditions, as seen by a high \tpr, high success in finding goals, but low success in reaching them. Overall, we see a positive affect when using both goal-conditioning and our active learning method.

Model precondition accuracy is not necessarily indicative of high performance. As shown in the first row of Fig.~\ref{fig:goalconditioning}, although \ra has both a high TNR and high TPR, it never finds plans to the goal because of insufficient task-relevant data. Additional results on the risk-tolerance schedule and candidate trajectory diversity can be found on our \href{https://sites.google.com/view/active-mde}{website}. 
\subsection{Effect of acquisition function on performance}
We analyze the impact of the aggregation function and discount factor on performance. We test $\gamma=0.9$ and compare to no discounting with $\gamma=1$. 
We observe an slight improvement in performance in both \gridworldname and \watersimenvname when using $\gamma=0.9$ and $h_{\mathrm{max}}$, which is consistent with our intuition that discounting can account for the dependence of later states on reaching earlier states. The improvement is smaller for \gridworldname and more apparent in the success rate in finding plans. 
Overall, we find that the choice of $\gamma$ and $h$ does not have a major impact on performance, but there may be some benefit in using a discount factor or other forms of aggregation of step-wise acquisition function values.


\section{Limitations}
\ed{8}{Our proposed active learning approach is limited to scenarios where dynamic variables that significantly affect deviation are represented in the state. Unobserved dynamic variables are implicitly modelled as noise, which may lead to overly restrictive model preconditions. This issue can be mitigated by prioritizing recent data or by incorporating these variables into the state, if feasible.} \ed{9}{Our MDE implementation with a GP does not directly scale to high-dimensional state spaces. Future work will explore using pre-trained general-purpose models to address this, potentially using cross-task information for added efficiency.}
 \\
 \\
 \\

\section{Conclusions}
This paper formulates the problem of active learning of model preconditions then presents a novel class of techniques designed to generate and select candidate trajectories. 
 We evaluate the performance of variations on our active learning approach on two simulated scenarios and one real-world task with learned models and analytical models. 
 Our experimental results demonstrate the effect of algorithmic choices in candidate trajectory selection and acquisition function on data efficiency. 
This work enables empirical estimation of model preconditions with minimal data, a capability we plan to extend to high-dimensional deformable object scenarios where the use of model preconditions can be particularly beneficial.
\\
\\
\label{sec:conclusion}

\bibliographystyle{ieeetr}
\bibliography{references} 
\end{document}